\documentclass[final]{cvpr}

\usepackage{times}
\usepackage{epsfig}
\usepackage{graphicx}
\usepackage{amsmath}
\usepackage{amssymb}
\usepackage{blindtext}
\usepackage{caption}
\usepackage{subcaption}
\usepackage{enumitem}

\usepackage[pagebackref=true,breaklinks=true,colorlinks,bookmarks=false]{hyperref}

\title{Multi-Label Activity Recognition using Activity-specific Features and Activity Correlations}
\author{Yanyi Zhang\\
Rutgers University–New Brunswick\\
\and
Xinyu Li\\
Amazon Web Service\\
\and
Ivan Marsic\\
Rutgers University–New Brunswick\\

}

\begin{document}
\maketitle

\begin{abstract}
Multi-label activity recognition is designed for recognizing multiple activities that are performed simultaneously or sequentially in each video. Most recent activity recognition networks focus on single-activities, that assume only one activity in each video. These networks extract shared features for all the activities, which are not designed for multi-label activities. We introduce an approach to multi-label activity recognition that extracts independent feature descriptors for each activity and learns activity correlations. This structure can be trained end-to-end and plugged into any existing network structures for video classification. Our method outperformed state-of-the-art approaches on four multi-label activity recognition datasets. To better understand the activity-specific features that the system generated, we visualized these activity-specific features in the Charades dataset.
\end{abstract}

\section{Introduction}

Activity recognition has been studied in recent years due to its great potential in real-world applications. Recent activity recognition researches \cite{kay2017kinetics, soomro2012ucf101, goyal2017something, kuehne2011hmdb} focused on single-activity recognition assuming that each video contains only one activity, without considering a multi-label problem where a video may contain multiple activities (concurrent or sequential). Multi-label activity recognition is an understudied field but has more general real-world use cases (e.g., sports activity recognition \cite{sozykin2018multi, carbonneau2015real}, or daily life activity recognition \cite{sigurdsson2016hollywood}). Most of the recent multi-label activity recognition methods are derived from structures for single activities that generate a shared feature vector and apply sigmoid as the output activation function \cite{li2017concurrent, wang2016temporal, carreira2017quo, wang2018non, feichtenhofer2019slowfast, wu2019long}. Although these approaches enable the network to provide multi-label outputs, the features are not designed for multi-label activities. 

\begin{figure}[!t]
	\includegraphics[width=0.48\textwidth]{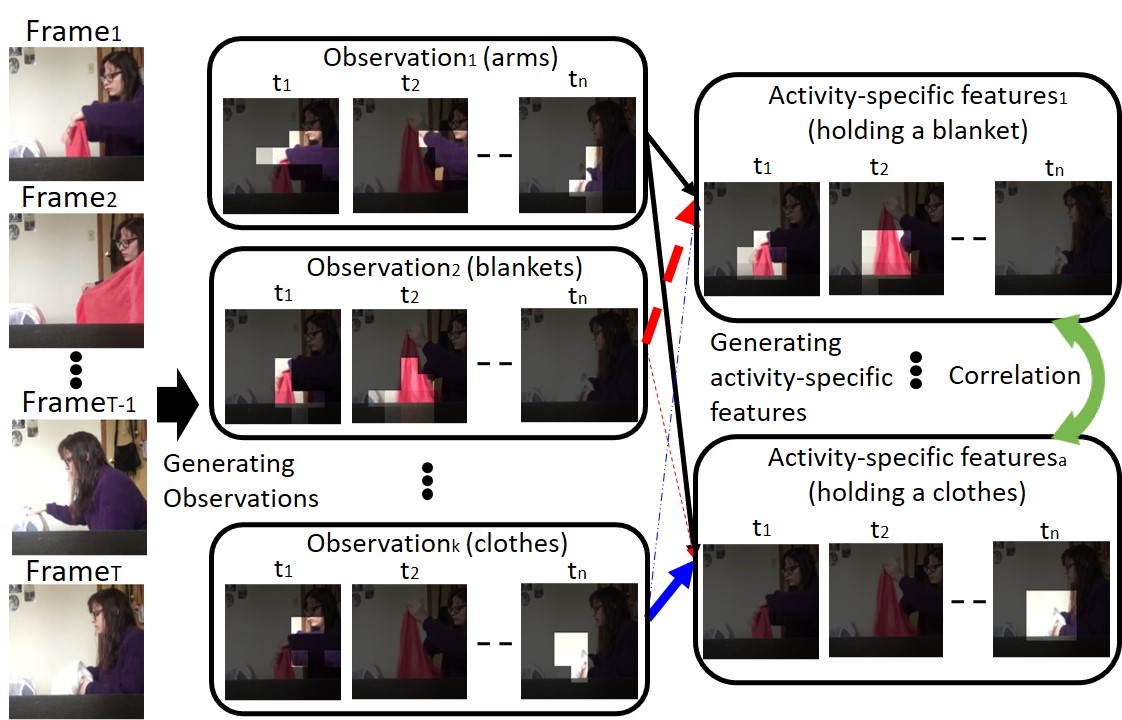}
	\vspace{-1.5em}
	\captionsetup{font=small}
	\caption{System overview using an example from Charades. The system first generates $k$ independent feature snippets (``observations'') that focus on different key regions from the video (arms, blankets, and clothes). The activity-specific features are then generated by independently combining these observations. The weights of the observations that contribute to activity-specific features are represented as lines with different colors (black, red, and blue). The thicker lines denote higher weights. For example, the $activity\textrm{-}specific \  features_1$ (holding a blanket) are obtained by combining information from $observation_1$ (focuses on arms) and $observation_2$ (focuses on clothes). The system finally learns correlations between activity-specific features and provide multi-label activity predictions. }
	\label{fig:01}
	\vspace{-2.0em}
\end{figure}

We introduce our mechanism to recognize multi-label activities from another angle by generating independent feature descriptors for different activities. We named these feature descriptors \textbf{``activity-specific features''}. This mechanism generates activity-specific features in two stages. The first-stage network (Figure \ref{fig:01}, middle) summarizes the feature maps extracted by the backbone network (3D convolution layers) and generates a set of independent feature snippets by applying independent spatio-temporal attention for each snippet. We name these feature snippets \textbf{``observations''}. In the second stage (Figure \ref{fig:01}, right), the network learns activity-specific features from different combinations of observations for different activities. In this way, each activity is represented as an independent set of feature descriptors (activity-specific features). The multi-label activity predictions can then be made based on the activity-specific features. Unlike most of the previous approaches \cite{li2017concurrent, wang2016temporal, carreira2017quo, wang2018non, feichtenhofer2019slowfast, wu2019long} that generate a shared feature vector to represent multiple activities by pooling feature maps globally, our network produces specific feature descriptors for each activity.

Label dependencies have proven important for multi-label image classification \cite{huynh2020interactive, guo2011multi, wang2016cnn}, and we argue that it is also important to consider the label-wise correlation for activities. We generate an activity label-wise \textbf{correlation map} to model co-existing patterns (e.g., walking and talking) and exclusive patterns (e.g., sitting and standing). The correlation map is applied on the activity-specific features to predict activities based on the highly correlated (or exclusive) activities' descriptors. In multi-label activity videos, different activities might have different duration and need to be recognized using video clips with different lengths. To address this issue, we further introduced a \textbf{speed-invariant tuning} method for generating activity-specific features and recognizing multi-label activities using inputs with different downsampling rates.

We evaluated our model on both large-scale multi-label activity datasets (Charades \cite{sigurdsson2016hollywood} and  AVA \cite{gu2018ava}), and real-world multi-label sports datasets (Volleyball \cite{sozykin2018multi} and Hockey \cite{carbonneau2015real}) to show that our model performs well on multi-label datasets and is applicable to real-world tasks. Our introduced model outperformed the recent state-of-the-art networks on all four datasets without using additional information (e.g., optical flow \cite{horn1981determining} or object information \cite{ji2020action}) other than RGB frames, which demonstrated the efficiency of our introduced method. We also provide detailed ablation experiments on the model structure to show that the introduced activity-specific features and activity correlation work as expected. We further visualized the activity-specific features by applying the learned attention maps on the backbone features (feature maps after the last 3D convolution layer) to represent the activity-specific feature maps. Our contributions can be summarized as:
\begin{itemize} [itemsep=0pt,parsep=0pt]
\item A network structure that generates activity-specific features for multi-label activity recognition. 
\item An activity correlation map that learns correlations between different activity-specific features.
\item A speed-invariant tuning method that produces multi-label activity predictions using different temporal-resolution inputs.
\end{itemize}

\begin{figure*}[ht]
\includegraphics[width=1\textwidth]{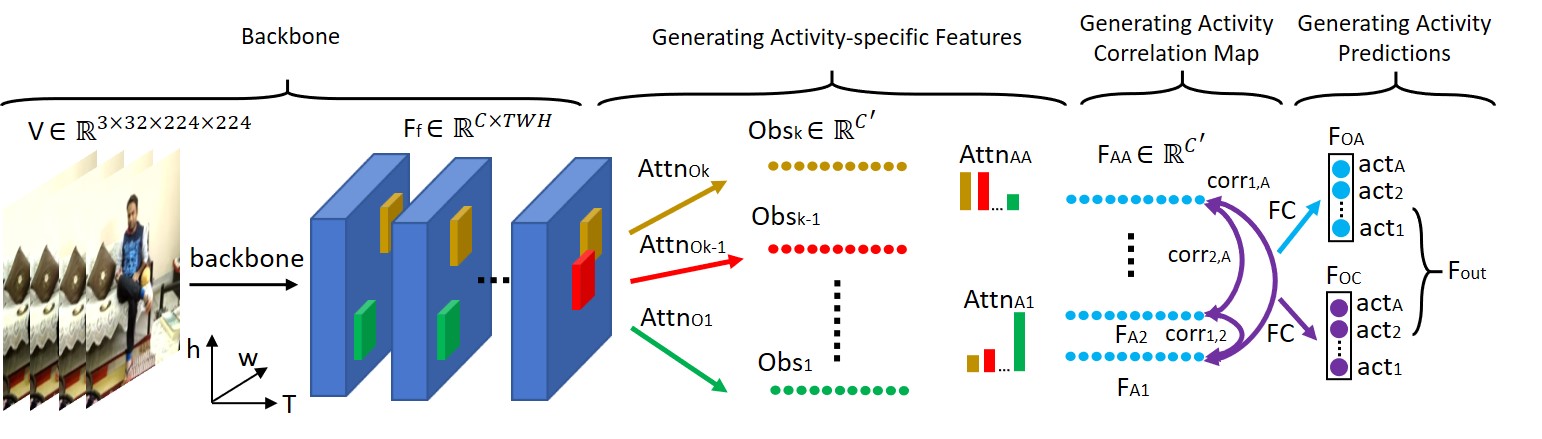}
\vspace{-2.6em}
\captionsetup{font=small}
\caption{Method overview, showing the detail dimension transformation when generating activity-specific features and providing predictions. Attention (red, green, and brown) focus on different spatio-temporal regions of the backbone feature ($F_f$) for generating observations ($Obs$), and generating activity-specific features ($F_A$) by combining observations using $Attn_A$. The purple lines ($corr_{j,k}$) are the correlations between pairs of activity-specific features.}
\label{fig:02}
\vspace{-1.5em}
\end{figure*}

\section{Related Work}
\textbf{Activity Recognition.} Video-based activity recognition has been developing rapidly in recent years due to the success of deep learning methods for image recognition \cite{krizhevsky2012imagenet, szegedy2015going, he2016deep}. Compared to image classification, activity recognition depends on spatio-temporal features extracted from consecutive frames instead of spatial-only features from static images. Two-stream networks apply two-branch convolution layers to extract motion features from consecutive frames as well as spatial features from static images and fuse them for activity recognition \cite{simonyan2014two, feichtenhofer2016convolutional, wang2016temporal}. Others proposed 3D-convolution-based networks for extracting spatio-temporal features from the videos instead of using manually designed optical flow for extracting motions between frames \cite{carreira2017quo, tran2015learning}. The nonlocal neural network \cite{wang2018non} and the long-term feature bank (LFB) \cite{wu2019long} extended the 3D ConvNet by extracting long-range features. The SlowFast network \cite{feichtenhofer2019slowfast} introduced a two-pathway network for learning motion and spatial features separately from the videos. The most recent X3D \cite{feichtenhofer2020x3d} and AssembleNet \cite{ryoo2019assemblenet, ryoo2020assemblenet++} build a searching network to find the best multi-branch architecture for the target dataset.

\textbf{Multi-label Activity Recognition.} Multi-label activity recognition is designed for recognizing multiple activities that are performed simultaneously or sequentially in each video \cite{sigurdsson2016hollywood, yeung2018every, ray2018scenes}. Most of the video networks focus on single-activity recognition and use sigmoid as the output activation function to provide multi-label predictions~\cite{carreira2017quo, wang2018non, feichtenhofer2019slowfast}. Other systems use weak-supervised learning and Zero-shot learning methods to help recognize weakly labeled activities \cite{mosabbeb2014multi, wang2020multi}. We introduce our network that improves the performance of multi-label activities at the feature-level by extracting independent feature descriptors for each activity instead of shared feature representations.

\textbf{Multi-label Image Classification.} Multi-label image classification has been studied for several decades. Previous approaches using graphical models to represent the label-wise dependency, such as Dependency Network \cite{guo2011multi}, and Correlative Model \cite{xue2011correlative}. These networks are not end-to-end trainable and require to input the entire dataset into the model, which is not generalizable to large-scale datasets. Others proposed end-to-end CNN-RNN architectures to model label-wise correlations \cite{chen2017order, wang2016cnn, yazici2020orderless}. These networks are still using shared features repeated for predicting each label, and RNN-based networks are too slow for training large-scale datasets, such as video datasets \cite{kay2017kinetics, sigurdsson2016hollywood}. 

\section{Methodology}
Recent state-of-the-art networks recognize activities by learning a shared feature representation for all the activities. These features are designed for recognizing single activities instead of multi-label activities, which requires to extract features for multiple activities in each video as well as the activity correlations. Our network focuses on multi-label activity recognition from a different angle: that of learning independent representations for each activity (activity-specific features) as well as their correlations (activity correlation map). Given a video clip $V \in \mathbb{R}^{3\times 32 \times 224 \times 224}$ with 32 consecutive frames, our model provides activity predictions in three steps:
\begin{enumerate} 
\item \textbf{Generating activity-specific features}:
we generate independent feature representations for $A$ different activities. This step consists of two sub-steps: we first generate $K$ spatio-temporally independent feature snippets (observations), $Obs \in \mathbb{R}^{K\times C'}$, that focus on different spatio-temporal regions of the video. We then apply attention $Attn_A$ on the observations to generate feature descriptors $F_A$ (activity-specific features) that are independent for each activity using independent weighted combinations of observations (Figure \ref{fig:02}, Generating Activity-specific Features).
\item \textbf{Generating activity correlation map}: we then generate an activity label-wise correlation map to represent the correlation between pairs of activities (Figure \ref{fig:02}, Generating Activity Correlation Map).
\item \textbf{Generating activity predictions}: we finally provide multi-label activity predictions use their corresponding activity-specific features and the features of correlated activities (Figure \ref{fig:02}, Generating Activity Predictions).
\end{enumerate}

\subsection{Generating Activity-specific Features}
Given a feature set $F_{f}\in \mathbb{R}^{C \times TWH}$ from the backbone network (e.g., i3D \cite{carreira2017quo}), the activity-specific features can be generated as:
\begin{equation} 
    F_A = \{Attn_{A1} Obs, Attn_{A2} Obs, ..., Attn_{AA} Obs\}
\end{equation}
where $F_A \in \mathbb{R}^{A \times C'}$ denotes $A$ independent feature descriptors for their corresponding activities (activity-specific features), $C'$ is the channel number of $F_A$, $Obs \in \mathbb{R}^{K\times C'}$ denotes $K$ independent feature snippets (observations) that are extracted from the backbone features $F_{f}$, and $Attn_{Ai}$ ($i \in {1, 2, ..., A}$) are the attentions that independently combine the $K$ observations to generate activity-specific features for the $i^{th}$ activity. We create these observations instead of directly generating $Attn_A$ from the backbone features $F_{f}$ to reduce redundant information. Each observation is an independent spatio-temporal feature snippet that focuses on a specific key region in a video. The $Obs$ are generated by applying $K$ independent spatio-temporal attentions on $F_{f}$ as:
\begin{equation}
    Obs_k= Attn_{Ok} [g_k^\alpha (F_{f})]^T
\end{equation}
\begin{equation}
    Obs = \{Obs_1, Obs_2, ..., Obs_k\}
\end{equation}
where $Obs_k \in \mathbb{R}^{C'}$ is the $k^{th}$ observation that focuses on a specific key region of the video, $Attn_{Ok} \in \mathbb{R}^{TWH}$ denotes the spatio-temporal attention for generating the $k^{th}$ observation. The method for generating $Attn_A$ and $Attn_O$ will be introduced later. The $g_k^\alpha$ is the linear function to integrate channels from $F_{f}$, which is represented as:
\begin{equation}
    \label{equ_linear}
    g_k^\alpha (F_{f}) = W_k^\alpha F_{f}
\end{equation}
where $W_k^\alpha \in \mathbb{R}^{C' \times C}$ are the weights for the linear function $g_k^\alpha$. The activity-specific set $F_A$ can finally be written as:
\begin{equation}
\begin{aligned}
     F_A = \{&Attn_{A1} \{Attn_{O1} [g_1^\alpha (F_{f}) ]^T, ..., Attn_{Ok} [g_k^\alpha (F_{f})]^T\}\\
     &Attn_{A2} \{Attn_{O1} [g_1^\alpha (F_{f}) ]^T, ..., Attn_{Ok} [g_k^\alpha (F_{f})]^T\}\\
     &\;\;\;\;\;\;\;\;\;\;\;\;\;\;\;\;\;\;\;\;\;\;\;\;\;\;\;\;\;\;\;\;...\\
     &Attn_{AA} \{Attn_{O1} [g_1^\alpha (F_{f}) ]^T, ..., Attn_{Ok} [g_k^\alpha (F_{f})]^T\}\}
\end{aligned}
\end{equation}


\subsection{Generating Activity Correlation Maps}
Label-wise correlations have proven important to be learned for multi-label classes \cite{xue2011correlative}. Representing co-existing and exclusive patterns between activities is also necessary for recognizing multi-label activities. We apply an activity label-wise correlation map onto the activity-specific feature set $F_A$ as:
\begin{equation}
    F_{AC} = Corr F_A
\end{equation}
where $F_{AC} \in \mathbb{R}^{A \times C'}$ is the result of matrix multiplication between $Corr$ and $F_A$. We applied the matrix multiplication between $Corr$ and $F_A$ to help recognize activities using their correlated activity-specific features. $Corr \in \mathbb{R}^{A \times A}$ is the correlation map generated as:
\begin{equation}
    Corr = Attn_C + Mask
\end{equation}
\begin{equation}
    Mask(j, k) = \frac{N_{j,k}}{N_j} 
\end{equation}
The correlation map $Corr$ is composed from $Mask \in \mathbb{R}^{A \times A}$ and $Attn_C \in \mathbb{R}^{A \times A}$, $Mask(j, k)$ is the frequency of activity $j$ co-existing with activity $k$ in the training ground truth, $N_{i,j}$ is the number of samples that includes both activity $j$ and $k$, and $N_{j}$ is the number of samples in which activity $j$ occurs. $Attn_C$ is the attention generated from $F_A$ to adjust the correlation map based on given different inputs. The method for generating $Attn_C$ is similar to $Attn_A$ and $Attn_O$, which will be introduced later.

\subsection{Generating Activity Predictions}
\label{section:predictions}
We finally make activity predictions using their corresponding activity-specific features and the features of correlated activities. To ensure that the $F_A$ represents activity-specific features, without mixing features of other activities from the loss propagated by $F_{AC}$, we build a multi-output network that predicts activities from both $F_A$ and $F_{AC}$ as: 
\begin{equation}
    F_{OA}=sigmoid(W^\varphi F_A+b^\varphi)
\end{equation}
\begin{equation}
    F_{OC}=sigmoid(W^\theta F_{AC}+b^\theta)
\end{equation}
\begin{equation}
    F_{out}=[F_{OA}, F_{OC}]
\end{equation}
where $F_{out} \in \mathbb{R}^{A \times 1}$ is the final output of the network with $F_{OA}$ and $F_{OC}$, two outputs for activity predictions. $W^\varphi$, $W^\theta$, $b^\varphi$, and $b^\theta$ are the weights and bias for predicting activities from $F_A$ and $F_{AC}$.

\subsection{Generating Attentions}
The attention mechanism was introduced for capturing long-term dependencies within sequential inputs and is commonly used in natural language processing systems \cite{vaswani2017attention, shen2017disan}. We applied $Attn_O$ and $Attn_A$ for generating activity-specific features, and $Attn_C$ for generating activity correlations. We implemented the dot-product attention method \cite{vaswani2017attention} for generating $Attn_O$ as:
\begin{equation}
\label{equ:attn}
    Attn_{Ok}=softmax([g_k^\beta(F_{f})_{-1}]^T g_k^\gamma (F_{f}))
\end{equation}
where $Attn_{Ok} \in \mathbb{R}^{TWH}$ denotes the attention for the $k^{th}$ observation, $g_k^\beta$, $g_k^\gamma$ are the linear functions same as the $g_k^\alpha$ in equation \ref{equ_linear}, and $g_k^\beta(F_{f})_{-1} \in \mathbb{R}^{C' \times 1}$ denotes selecting the last row of the  $g_k^\beta(F_{f}) \in \mathbb{R}^{C' \times TWH}$ to produce an appropriate dimension for $Attn_{Ok}$. The $Attn_A$ and $Attn_C$ were generated using a similar approach. Other attention methods (e.g., additive attention \cite{bahdanau2014neural}) could be used for generating attentions, but we selected the dot-product attention method because previous research has shown that it is more efficient and works well for machine translation \cite{vaswani2017attention, shen2017disan}.

Applying linear functions in equation (\ref{equ_linear}) requires a large number of weights. Inspired by the channel-separated network \cite{tran2019video}, we applied a group linear function to reduce the number of weights for $g_k^\alpha$ by splitting $F_{f}$ into $n$ groups based on the channel dimension and applying independent linear functions to each group as:
\begin{equation}
    \label{eq_conv1d}
    g_k^\alpha(F_{f}) = Concat(W_{k1}^{\alpha}F_{f1}, W_{k2}^{\alpha}F_{f2}, ..., W_{kn}^{\alpha}F_{fn})
\end{equation}
where $F_{fn} \in \mathbb{R}^{\frac{C}{n} \times TWH}$ is the $n^{th}$ group of $F_f$, and $W_{kn} \in \mathbb{R}^{\frac{C'}{n} \times \frac{C}{n}}$ are the weights of the linear function for $F_{fn}$. Using a larger number of groups ($n$) results in fewer parameters ($n$ times fewer). We set $n=32$ empirically to minimize the number of weights without affecting the model performance.

\subsection{Comparison with Attention-based Networks}
We compared our network with other attention-based networks to show two differences \cite{girdhar2017attentional, du2017recurrent, li2018unified, meng2019interpretable, girdhar2017attentional}. First, previous approaches apply attentions shared for all the channels in $F_{f}$ and generate shared feature descriptors for all the activities. In our approach, observations are generated using independent attentions and focus on different key regions of the video. Second, attentions in previous methods only work through spatio-temporal dimensions, while our $Attn_A$ works through observations to generate activity-specific features. In addition, our network also learns activity correlations, which is not included in the other attention-based methods. We did an ablation experiment (Sect. \ref{sect:ablation}) to compare our network to the two-stage transformer and found that our network significantly outperformed the two-stage transformer.

\begin{figure}[!t]
	\includegraphics[width=0.48\textwidth]{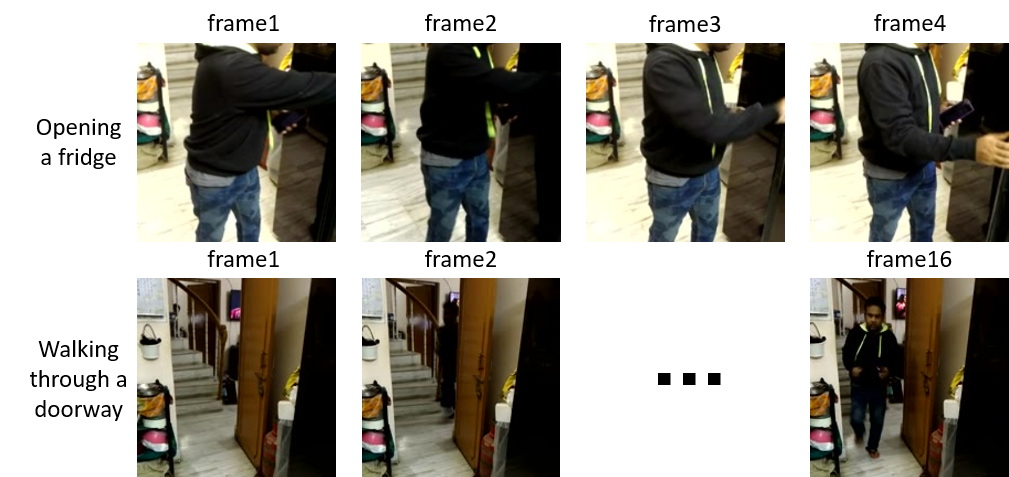}
	\vspace{-1.5em}
	\captionsetup{font=small}
	\caption{A video example in Charades shows multiple activities (opening a fridge and walking through a doorway) that require inputs with different sample rates.}
	\label{fig:multi-scale}
	\vspace{-1.5em}
\end{figure}

\subsection{Speed-invariant Tuning}
In multi-label activity videos, different activities may have different duration in a large range compared with single activity videos. Figure \ref{fig:multi-scale} shows an example in Charades having two activities: opening a fridge (short) and walking through a doorway (long). The LFB method \cite{wu2019long} proved that combining long-term features helps to improve the performance of recognizing multi-label activities. The LFB method requires extra FLOPs, and is not end-to-end trainable. We introduced our speed-invariant tuning method that ensures activities of different durations are properly covered in fixed frame inputs (32-frame) without requiring extra FLOPs.\\
\textbf{Training}: We first trained the complete model using the downsampling rate of 4 and froze the weights for all the 3D convolution layers. We then started finetuning the module after the 3D convolutions for $I$ iterations by using 32-frame inputs obtained by randomly selecting downsampling rate $r$ among 2, 4, and 8.\\
\textbf{Inference:} During the evaluation stage, we applied the 30-view test (followed \cite{feichtenhofer2019slowfast}) by selecting views from different downsampling rates (12 from $r=4$, and 9 from both $r=8$ and $r=2$) and summed the model predictions using inputs from the 30 views for the video-level activity prediction. We set the initial sampling rate $r$ to 4 following previous work \cite{wang2018non}. The model can then recognize activities that have different duration by aggregating the results from branches that used different downsampling rates as the input.

We noticed that the previous multi-grid algorithm \cite{wu2020multigrid} has a similar idea to speed up the model training process. Unlike the multi-grid algorithm, we merged the predictions from multiple input speeds for the final prediction (without additional parameters and FLOPs) instead of making the model converge to a specific shape at later epochs. Our speed-invariant method had the performance improved based on the model trained with fixed-speed inputs, while the multi-gird method achieved similar performance as the baseline. 

\subsection{Implementation Details}
We implemented our model with PyTorch \cite{paszke2017automatic}. We used batch normalization \cite{ioffe2015batch} and ReLU activation \cite{hahnloser2001permitted} for all the convolution layers. We used binary cross-entropy loss and the SGD optimizer with an initial learning rate $3.5e-2$ and $1.25e-5$ as the weight decay. Dropout (rate=0.5) was used after the dense layer to avoid overfitting \cite{srivastava2014dropout}. We set the batch size to 12 and trained our model with 3 RTX 2080 Ti GPUs for 50k iterations. We applied the scale-jittering method in the range of [256, 320] and horizontal flipping to augment the frames \cite{feichtenhofer2019slowfast}.  During the evaluation, we used the 30-view test \cite{feichtenhofer2019slowfast}.

\begin{table}
\captionsetup{font=small}
\caption{Comparison with other networks on Charades using a single RGB branch. The evaluation metric is mAP (mean-average-precision) in percentages, calculated using the officially provided script. The Slowfast* and Nonlocal* are the results of the baseline that we trained on Charades instead of the one reported in paper~\cite{feichtenhofer2019slowfast, wang2018non}.}
\vspace{-2.0em}
\label{table:01}
\small
\begin{center}
\begin{tabular}{|l|c|c|c|}
\hline
method & backbone &pre-train &mAP \\
\hline\hline
2D CNN \cite{sigurdsson2016hollywood} & Alexnet  & ImageNet & 11.2\\
MultiScale TRN \cite{zhou2018temporal} & Inception& ImageNet &25.2\\
\hline
I3D \cite{carreira2017quo} & Inception& Kinetics-400& 32.9\\
STRG \cite{wang2018videos} & Nonlocal-101& Kinetics-400 & 39.7\\
LFB \cite{wu2019long} & Nonlocal-101& Kinetics-400 & 42.5\\
Multi-Grid \cite{wu2020multigrid} & SlowFast-50& Kinetics-400 & 38.2\\
Nonlocal \cite{wang2018non} & Nonlocal-101 & Kinetics-400 & 37.5\\
Nonlocal* (Baseline) & Nonlocal-101 & Kinetics-400 & 40.3\\
\textbf{Our's} & Nonlocal-101 & Kinetics-400 & \textbf{44.2}\\
\hline
SlowFast \cite{feichtenhofer2019slowfast} & SlowFast-101& Kinetics-600  & 45.2\\
X3D-XL \cite{feichtenhofer2020x3d} & X3D-XL& Kinetics-600 & 47.2\\
Slowfast* (Baseline) & SlowFast-101& Kinetics-600 & 44.7\\
\textbf{Our's} & SlowFast-101& Kinetics-600 & \textbf{48.1}\\
\hline
CSN \cite{tran2019video} (Baseline) & CSN-152& IG-65M & 46.4\\
\textbf{Our's} & CSN-152& IG-65M & \textbf{50.3}\\
\hline
\end{tabular}
\end{center}
\vspace{-2.5em}
\end{table}

\begin{table}
\captionsetup{font=small}
\caption{Comparison with state-of-art networks on Charades using ensemble and additional inputs. The ens. denotes the network use ensemble approach to improve the performance.}
\vspace{-2.0em}
\label{table:additional}
\small
\begin{center}
\begin{tabular}{|l|c|c|c|}
\hline
method & extra inputs &mAP &\#backbones\\
\hline\hline
Hallucinating \cite{wang2019hallucinating} & IDT+Flow & 43.1& 1\\
AssembleNet ++ \cite{girdhar2017actionvlad} & Flow+Object& 59.9& 4\\
Action-Genome \cite{ji2020action} & Object+Person & 60.1 & 1\\
\hline
CSN \cite{tran2019video} (Baseline) & Object & 63.1& 1\\
\textbf{Our's} & Object & \textbf{65.5}& 1\\
\hline
\end{tabular}
\end{center}
\vspace{-3.3em}
\end{table}

\begin{table*}[t]
\small
\centering
\captionsetup{font=small}
\caption{Ablation experiments on the Charades dataset. We show the mAP scores, parameter numbers, and GFLOPs by using different hyper-parameters, backbone networks, and removing different modalities from our network.}
\vspace{-0.3cm}
 \begin{subtable}[t]{0.35\textwidth}
    \begin{tabular}[t]{|p{0.25\textwidth}|p{0.13\textwidth}|p{0.15\textwidth}|p{0.25\textwidth}|}
    \hline
    group size & mAP & Params & GFLOPs\\
    \hline
    \hline
    1 & 47.9 & 50.4M & 9.87 $\times$ 30\\
    8 & 48.3 & 6.3M& 1.24 $\times$ 30\\
    \textbf{32} & \textbf{48.4} & 1.6M & 0.32 $\times$ 30\\
    64 & 47.2 & 0.8M & 0.16 $\times$ 30\\
    \hline
    \end{tabular}
    \subcaption{Group size: performance on Charades using different group sizes by setting observation number to 64 and sample rate to 4.}
    \label{table:02.a}
  \end{subtable}
  \hspace{0.5cm}
  \begin{subtable}[t]{0.32\textwidth}
   \centering
      \begin{tabular}[t]{|p{0.22\textwidth}|p{0.13\textwidth}|p{0.15\textwidth}|p{0.25\textwidth}|}
        \hline
        obs num &mAP &Params& GFLOPs\\
        \hline
        \hline
        16 & 46.1 & 0.4M& 0.08 $\times$ 30\\
        32 & 47.5 & 0.8M& 0.16 $\times$ 30\\
        \textbf{64} & \textbf{48.4} & 1.6M& 0.32 $\times$ 30\\
        128 & 48.4 & 3.2M & 0.62 $\times$ 30\\
        \hline
        \end{tabular}
        \subcaption{Observation number: performance on Charades using different number of observations by setting group size to 32 and sample rate to 4.}
        \label{table:02.b}
   \end{subtable}
   \hspace{0.5cm}
   \begin{subtable}[t]{0.25\textwidth}
   \centering
      \begin{tabular}[t]{|p{0.7\textwidth}|p{0.15\textwidth}|}        
        \hline
        model structures &mAP\\
        \hline
        \hline
        baseline & 46.4\\
        no activity-specific & 46.6\\
        no activity correlation & 48.4\\
        \textbf{complete} & \textbf{49.2}\\
        \hline
        \end{tabular}
        \subcaption{Model structure ablation: performance on Charades after removing each module.}
        \label{table:02.d}
   \end{subtable}
   \hspace{0.5cm}
  \begin{subtable}[t]{0.5\textwidth}
   \centering
      \begin{tabular}{|p{0.2\textwidth}|p{0.1\textwidth}|p{0.5\textwidth}|}        
        \hline
        sample rate &mAP &GFLOPs\\
        \hline
        \hline
        4 only & 49.2 & (backbone + 0.32) $\times$ 30\\
        2 + 4 & 49.9 & (backbone + 0.32) $\times$ (15 + 15)\\
        \textbf{2 + 4 + 8} & \textbf{50.3} & (backbone + 0.32) $\times$ (9 + 12 + 9)\\
        \hline
        \end{tabular}
        \subcaption{Sample rate for speed-invariant tuning: performance and FLOPs of the model when using different sample rates by setting observation number to 64 and group size to 16.}
        \label{table:02.c}
   \end{subtable}
   \hspace{0.5cm}
 \hspace{0.5cm}
  \begin{subtable}[t]{0.27\textwidth}
    \centering
      \begin{tabular}{|p{0.65\textwidth}|p{0.15\textwidth}|}        
        \hline
        method & mAP\\
        \hline
        \hline
        Baseline & 46.4\\
        Two-Stage Transformer & 46.3\\
        \textbf{Our's} & \textbf{48.4}\\
        \hline
        \end{tabular}
        \subcaption{Two-stage transformer: compare the performance on Charades between our network and the two-stage transformer.}
   \label{table:02.f}
 \end{subtable}
\label{table:02}
\vspace{-2.5em}
\end{table*}

\section{Experiments on Charades}
Charades dataset \cite{sigurdsson2016hollywood} contains 9848 videos with average length of 30 seconds. This dataset includes 157 multi-label daily indoor activities. We used the officially provided train-validate split (7985/1863) to evaluate the network. We used the officially-provided 24-fps RGB frames as input and the officially-provided evaluation script for evaluating the validation set.

\subsection{Single RGB Branch Results}
We first compared our network with other state-of-art networks on Charades using a single RBG branch (only RGB videos as input and no ensemble). We compared our system with three baseline networks, CSN \cite{tran2019video}, pre-trained on IG-65M \cite{ghadiyaram2019large}), Slowfast-101, pre-trained on Kinetics-600, and Nonlocal-101, pre-trained on Kinetics-400 as well as other state-of-the-art methods that work on Charades. Our method achieved around \textbf{4\%} higher mAP score on Charades over all the baseline networks pre-trained with different datasets. This shows that the generated activity-specific features and the activity label-wise correlations help the network to work better for multi-label activity recognition by plugging into any video networks. We also outperformed all the other methods \cite{wu2019long, hussein2019timeception, wang2018videos, feichtenhofer2019slowfast} including the recent state-of-the-art approach, X3D \cite{feichtenhofer2020x3d} that pre-trained on Kinetics-600 \cite{carreira2018short} on Charades. Because the model performance on Charades highly depends on the pre-trained backbone network, our method could be further improved if we could use the most recent X3D as backbone~\cite{feichtenhofer2020x3d}.

\subsection{Data Reinforced Model Results}
Previous research showed that using additional inputs (e.g., flows and object information) improves the performance on Charades. The recent AssembleNet++ \cite{ryoo2019assemblenet, ryoo2020assemblenet++} built a searching algorithm to fuse a 4-branch network that included optical flow and object information beside RGB frames as additional inputs. The Action-Genome builds a graph network to learn the relationship between objects and persons for activity prediction \cite{ji2020action}. For a fair comparison, we evaluated our network on Charades by including the object information provided by \cite{ji2020action} as inputs. Table \ref{table:additional} show that our network also outperformed the baseline and other state-of-the-art networks using additional inputs. These results demonstrate that the activity-specific features and activity correlations also help for multi-label activity recognition with additional inputs included.

\subsection{Ablation Experiments}
\label{sect:ablation}
We next ablated our system with various hyper-parameters (group size, observation number, and sampling rate for speed-invariant inputs).

\textbf{Group sizes.} Table \ref{table:02.a} shows the system performance for different values of the group size ($n$) in equation \ref{eq_conv1d} when generating observations (64 observations and the downsampling rate is 4). The performance on Charades stayed at 48\% when the group size increased from 1 to 32 but dropped quickly for $n=64$. A larger group size results in using a smaller subset of channels from $F_{f}$ for generating observations, which requires fewer parameters but may cause a performance drop because of information loss.

\textbf{Observations number.} Table \ref{table:02.b} compares the system performance for different numbers of observations (group size is 32 and the downsampling rate is 4). The best-performing number of observations is 64, which also requires the fewest weights. Using a larger number of observations helps cover more key parts from the videos but the performance saturates for more than 64 observations.

\begin{table}
\captionsetup{font=small}
\caption{Complexity analysis of our network. We compare the additional FLOPs and parameters that our network require to the other backbone networks.}
\vspace{-1.5em}
\label{table:flops}
\small
\begin{center}
\begin{tabular}{|l|c|c|}
\hline
method &Params &GFLOPs \\
\hline\hline
Nonlocal-101 \cite{sigurdsson2016hollywood}& 54.3 M & 359.0 $\times$ 30\\
Slowfast-101 \cite{girdhar2017actionvlad} &59.9 M & 234.0 $\times$ 30\\
CSN-152 \cite{tran2019video} & 32.8 M & 109.0 $\times$ 30\\
X3D-XL \cite{feichtenhofer2020x3d} & 11.0 M & 48.4 $\times$ 30\\
\hline
Our's-157 activities & 1.6 M & 0.32 $\times$ 30\\
Our's-1024 activities & 1.7 M & 0.33 $\times$ 30\\
\hline
\end{tabular}
\end{center}
\vspace{-4.0em}
\end{table}

\textbf{Model structure ablation.} We next evaluated the system on Charades by removing each component from our network. Table \ref{table:02.d} shows that the model without activity-specific features achieved similar performance as the baseline model although with the correlation map. Without learning activity-specific features, the system only learns correlations between different feature channels that have already have been learned in convolutions. The network with activity correlation achieved 1\% mAP score improvement on Charades that helped to predict co-existed activities and eliminate exclusive activities. These results show that our network achieved a performance boost by the combination of key components of our network.

\textbf{Sampling rates for speed-invariant tuning.} We evaluated our speed-invariant tuning method by merging predictions using different downsampling rates at inputs. Table \ref{table:02.c} shows that speed-invariant models achieved better performance compared to the model using a single downsampling rate of 4 because the speed-invariant tuning method makes the features better represent activities of different duration. The system achieved the best performance by merging predictions based on 2, 4, and 8 downsampling rates and this method did not require extra FLOPs by using the same view numbers (30 views) for evaluation (Table \ref{table:02.c} row 3).

\textbf{Two-stage transformer.} We then compared our network with the two-stage transformer (which might incorrectly appear the same as our network) that stack two self-attention layers after the backbone features to show the difference. Each self-attention layer is the multiplication between the attention generated using equation \ref{equ:attn} and the $F_{f}$. We then applied a 3D global average pooling after the two-stage self-attention to average the features over the spatio-temporal dimensions and a fully-connected layer to make predictions. The two-stage transformer makes all the channels share the same attention that also attempts to learn features shared by all the activities instead of activity-specific features. Table \ref{table:02.f} shows that the two-stage transformer did not increase the performance compared to the baseline network, while our network significantly outperformed the baseline network. 

\textbf{Complexity Analysis.} We finally calculated the additional complexity that our network required for learning activity-specific features. Table \ref{table:flops} shows that the additional parameters and FLOPs for our network are negligible compared to the complexity of the backbone networks. We also calculated the complexity of our network when the number of activities is 1024 (Table \ref{table:flops} last row). The complexity of our network is not linearly related to the activity number because of the design of observation generation to reduce redundant parameters. Our network only requires small amounts of additional computing resources based on the backbone networks even with a large number of activities.

\begin{figure*}[ht]
\includegraphics[width=1\textwidth]{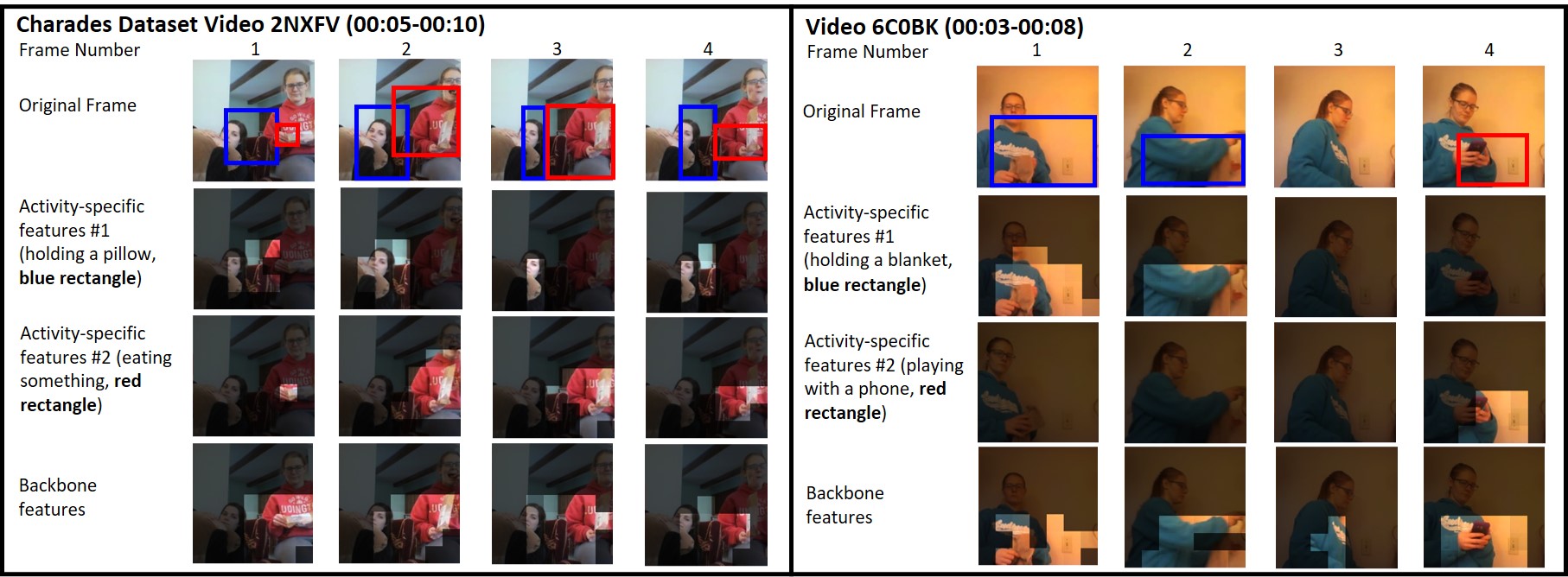}
\captionsetup{font=small}
\vspace{-1.7em}
\caption{Visualizing the activity-specific features in two videos from the Charades dataset. The bounding boxes in the original frames correspond to the activated regions in the activity-specific feature maps. The activity-specific feature maps will only focus on the regions where corresponding activities are being performed and have low values if there is no activity being performed at that time.}
\label{fig:03}
\vspace{-1.8em}
\end{figure*}

\begin{table}
\captionsetup{font=small}
\caption{Comparison with other networks on AVA. The evaluation metric is mean Average Precision using 0.5 IoU as threshold. The result of Slowfast is evaluated using official released parameters instead of the result reported in paper \cite{feichtenhofer2019slowfast}.}
\vspace{-2.0em}
\label{table:ava}
\small
\begin{center}
\begin{tabular}{|l|c|c|c|}
\hline
method &backbone &pre-train& mAP \\
\hline\hline
ATR \cite{jiang2018human} & Nonlocal-50& Kinetics-400& 21.7\\
LFB \cite{wu2019long} & Nonlocal-101& Kinetics-400& 27.7\\
Slowfast-101 \cite{feichtenhofer2019slowfast} &Slowfast-101& Kinetics-600& 29.1\\
X3D-XL \cite{feichtenhofer2020x3d} & X3D-XL& Kinetics-600& 27.4\\
\textbf{Our's} &Slowfast-101& Kinetics-600& \textbf{30.2}\\
\hline
CSN-152 (Baseline) & CSN-152 & IG-65M & 27.9\\
\textbf{Our's}& CSN-152& IG-65M & \textbf{29.2}\\
\hline
\end{tabular}
\end{center}
\vspace{-3.5em}
\end{table}

\section{Experiments on AVA}
We next evaluated our network on another large-scale dataset, AVA \cite{gu2018ava}, for multi-label activity detection. The AVA dataset includes 211k video clips for training and 57k clips for validating. Each clip has one keyframe with multiple persons localized using a bounding box and labeled with 80 multi-label activities. Following the official protocol \cite{gu2018ava}, we evaluated the network on 60 activities using mean Average Precision (mAP) score and 0.5 IoU as the threshold. To make a fair comparison, we used the person bounding boxes generated by previous work and threshold the detected persons having confidence score~$>$~0.8 \cite{feichtenhofer2019slowfast, feichtenhofer2019slowfastcode}. We applied the ROI-Align algorithm \cite{ren2015faster} to extract the features of the persons using their corresponding bounding boxes. We set the spatial stride of the last stage of the backbone from 2 to 1 to increase the spatial resolution ($14 \times 14$).

We compared our method with the baseline and other state-of-the-art networks on AVA in Table \ref{table:ava}. We again outperformed the current state-of-the-art network (Slowfast-101, officially released parameters \cite{feichtenhofer2019slowfastcode}) on AVA using center-crop views (instead of 6-crop, and multi-scale).  This demonstrated that our proposed activity-specific feature and activity correlation map generalizes well to different datasets and consistently improves multi-label classification problems by a significant margin.

\begin{table}[t]
\centering
\small
\captionsetup{font=small}
\caption{Experimental results on Volleyball. The ``s'' and ``bb'' in the last two columns denote using the whole scene and bounding boxes of persons as supplemental for recognizing group activities.}
\label{table:03}
\vspace{-1.0em}
\begin{tabular}{|l|c|cc|}
\hline
\multicolumn{2}{|c}{Volleyball Personal (multi-label)}&
\multicolumn{2}{c|}{Volleyball Group}\\
\hline\hline
method & Acc. & Acc. (s) & Acc. (bb)\\ \hline
Hier LSTM \cite{ibrahim2016hierarchical} & 72.7 & 63.1 &81.9\\
SRNN \cite{biswas2018structural} & 76.6 & -&83.4\\
So-Sce \cite{bagautdinov2017social} & 82.4 &75.5&89.9\\
CRM \cite{azar2019convolutional} & - &75.9&93.0\\
Act-trans \cite{gavrilyuk2020actor} & 85.9 &-&94.4\\
CSN-152 baseline & 85.0 &87.1&-\\
\textbf{Our's} & \textbf{86.6} &\textbf{87.6}&\textbf{95.5}\\
\hline
\end{tabular}
\vspace{-2.4em}
\end{table}

\section{Experiments on Hockey and Volleyball}
We also run experiments on two small datasets, Hockey and Volleyball \cite{ibrahim2016hierarchical,sozykin2018multi}. The Volleyball Dataset contains 55~videos with 4830 annotated video clips. This dataset includes two sets of labels for group activity recognition task (8-class multi-class classification) and multi-label activity recognition task (9-label multi-label classification). We evaluated our method on both of these tasks. \textbf{The experimental results on Hockey is in the supplemental material because of the page limit.}

Table \ref{table:03} shows that our system substantially outperformed all the existing approaches on Volleyball for multi-label activities \cite{gavrilyuk2020actor}. We also compared our method with the baseline model using the latest backbone network (CSN baseline in Table \ref{table:03}) that works on activity recognition. Our system achieved roughly 2\% higher accuracy score compared with the baseline model, which shows that the activity-specific features also improve multi-label activity recognition on small sports datasets.

We further evaluated our method on Volleyball for group activity recognition. The group activity is essentially a single-activity recognition problem: only one activity occurs during one video clip. Our method outperformed other state-of-the-art methods when using the whole scene (s, Table \ref{table:03}) as input \cite{gavrilyuk2020actor} (RGB frames without using bounding boxes around people). This shows that our method generalizes to the single-activity recognition problem as well. Previous methods \cite{azar2019convolutional, gavrilyuk2020actor} used bounding boxes around people (bb, Table \ref{table:03}) and their individual activities as supplemental information for group activity recognition. We tested our model by including this supplemental information, and our approach outperformed the recent state-of-the-art method \cite{azar2019convolutional} (95.5 for our system vs. 94.4 for the Act-trans in the last column of Table~\ref{table:03}). Compared to the baseline network, our method slightly outperformed the baseline network (87.2 vs. 87.1 in the second-to-the-last column of Table \ref{table:03}). The activity-specific features do not help significantly in the single-activity problems, unlike the case of multi-label activities, because the feature maps will only focus on one region where the single-activity occurred.

\section{Feature Visualization}
To better understand what activity-specific features are learned, we visualized these features for the activities present in the video clips. Figure \ref{fig:03} shows two examples in the Charades, including the activity-specific feature maps (last two rows of each example in Figure \ref{fig:03}) and their corresponding input frames. The activity-specific feature maps were generated by applying the learned $Attn_O$ and $Attn_A$ on the backbone features $F_{f}$. We normalized the feature maps between 0 and 1 and plotted these maps for the activities present in the video (last two rows of each example in Figure \ref{fig:03}). To make the visualized maps more understandable, we applied the 0.5 threshold to the activity-specific feature maps and drew the bounding boxes using different colors (blue, red) for different activities in the original frames around the regions activated in the feature maps.

Based on the visualizations in Figure \ref{fig:03}, we can make three points. First, unlike the backbone features (last row in Figure \ref{fig:03}) the activity-specific features will only focus on the spatial regions for the corresponding activity when multiple activities are performed simultaneously. The visualization of video ``2NXFV'' (Figure \ref{fig:03}, left) shows the activity-specific features \#1 (holding a pillow) focusing the region of the left person, and activity-specific features \#2 (eating something) focusing the right person who is performing the corresponding activity. Second, only the activity-specific features corresponding to the activity being performed will have high values when the video has one activity or activities performed in sequence. The visualization of video ``6C0BK'' (Figure \ref{fig:03}, right) shows the activity-specific features \#1 (holding a blanket) having activated regions at the first two frames, while the activity-specific features \#2 (playing with a phone) focused on the last frames. Finally, all the activity-specific features will have low values for the frames in which no activities were performed (Figure \ref{fig:03}, the third column of the right diagram). These visualizations demonstrate that the activity-specific features will focus on the key regions from the video that are related to their corresponding activities.

\section{Conclusion and Future Work}
We introduced a network that focuses on multi-label activity recognition. The network generates spatio-temporally independent activity-specific features for each activity and learns activity correlations. We outperformed previous state-of-the-art methods on four multi-label activity recognition datasets. The visualizations showed that the activity-specific features are representative of their corresponding activities. One issue remains in the speed-invariant tuning method, where we simply summed the predictions by using different downsampling rates for the inputs. Extending the speed-invariant method to enable the model to learn to select features from appropriate speeds for different activities will be our future work.

{\small
\bibliographystyle{ieee_fullname}
\bibliography{egbib}

\begin{thebibliography}{10}\itemsep=-1pt

\bibitem{azar2019convolutional}
Sina~Mokhtarzadeh Azar, Mina~Ghadimi Atigh, Ahmad Nickabadi, and Alexandre
  Alahi.
\newblock Convolutional relational machine for group activity recognition.
\newblock In {\em Proceedings of the IEEE Conference on Computer Vision and
  Pattern Recognition}, pages 7892--7901, 2019.

\bibitem{bagautdinov2017social}
Timur Bagautdinov, Alexandre Alahi, Fran{\c{c}}ois Fleuret, Pascal Fua, and
  Silvio Savarese.
\newblock Social scene understanding: End-to-end multi-person action
  localization and collective activity recognition.
\newblock In {\em Proceedings of the IEEE Conference on Computer Vision and
  Pattern Recognition}, pages 4315--4324, 2017.

\bibitem{bahdanau2014neural}
Dzmitry Bahdanau, Kyunghyun Cho, and Yoshua Bengio.
\newblock Neural machine translation by jointly learning to align and
  translate.
\newblock {\em arXiv preprint arXiv:1409.0473}, 2014.

\bibitem{biswas2018structural}
Sovan Biswas and Juergen Gall.
\newblock Structural recurrent neural network (srnn) for group activity
  analysis.
\newblock In {\em 2018 IEEE Winter Conference on Applications of Computer
  Vision (WACV)}, pages 1625--1632. IEEE, 2018.

\bibitem{carbonneau2015real}
Marc-Andr{\'e} Carbonneau, Alexandre~J Raymond, Eric Granger, and Ghyslain
  Gagnon.
\newblock Real-time visual play-break detection in sport events using a context
  descriptor.
\newblock In {\em 2015 IEEE International Symposium on Circuits and Systems
  (ISCAS)}, pages 2808--2811. IEEE, 2015.

\bibitem{carreira2018short}
Joao Carreira, Eric Noland, Andras Banki-Horvath, Chloe Hillier, and Andrew
  Zisserman.
\newblock A short note about kinetics-600.
\newblock {\em arXiv preprint arXiv:1808.01340}, 2018.

\bibitem{carreira2017quo}
Joao Carreira and Andrew Zisserman.
\newblock Quo vadis, action recognition? a new model and the kinetics dataset.
\newblock In {\em proceedings of the IEEE Conference on Computer Vision and
  Pattern Recognition}, pages 6299--6308, 2017.

\bibitem{chen2017order}
Shang-Fu Chen, Yi-Chen Chen, Chih-Kuan Yeh, and Yu-Chiang~Frank Wang.
\newblock Order-free rnn with visual attention for multi-label classification.
\newblock {\em arXiv preprint arXiv:1707.05495}, 2017.

\bibitem{du2017recurrent}
Wenbin Du, Yali Wang, and Yu Qiao.
\newblock Recurrent spatial-temporal attention network for action recognition
  in videos.
\newblock {\em IEEE Transactions on Image Processing}, 27(3):1347--1360, 2017.

\bibitem{feichtenhofer2020x3d}
Christoph Feichtenhofer.
\newblock X3d: Expanding architectures for efficient video recognition.
\newblock In {\em Proceedings of the IEEE/CVF Conference on Computer Vision and
  Pattern Recognition}, pages 203--213, 2020.

\bibitem{feichtenhofer2019slowfastcode}
Christoph Feichtenhofer and Haoqi Fan.
\newblock Slowfast released code, https://github.com/facebookresearch/slowfast.
\newblock 2019.

\bibitem{feichtenhofer2019slowfast}
Christoph Feichtenhofer, Haoqi Fan, Jitendra Malik, and Kaiming He.
\newblock Slowfast networks for video recognition.
\newblock In {\em Proceedings of the IEEE international conference on computer
  vision}, pages 6202--6211, 2019.

\bibitem{feichtenhofer2016convolutional}
Christoph Feichtenhofer, Axel Pinz, and Andrew Zisserman.
\newblock Convolutional two-stream network fusion for video action recognition.
\newblock In {\em Proceedings of the IEEE conference on computer vision and
  pattern recognition}, pages 1933--1941, 2016.

\bibitem{gavrilyuk2020actor}
Kirill Gavrilyuk, Ryan Sanford, Mehrsan Javan, and Cees~GM Snoek.
\newblock Actor-transformers for group activity recognition.
\newblock {\em arXiv preprint arXiv:2003.12737}, 2020.

\bibitem{ghadiyaram2019large}
Deepti Ghadiyaram, Du Tran, and Dhruv Mahajan.
\newblock Large-scale weakly-supervised pre-training for video action
  recognition.
\newblock In {\em Proceedings of the IEEE Conference on Computer Vision and
  Pattern Recognition}, pages 12046--12055, 2019.

\bibitem{girdhar2017attentional}
Rohit Girdhar and Deva Ramanan.
\newblock Attentional pooling for action recognition.
\newblock In {\em Advances in Neural Information Processing Systems}, pages
  34--45, 2017.

\bibitem{girdhar2017actionvlad}
Rohit Girdhar, Deva Ramanan, Abhinav Gupta, Josef Sivic, and Bryan Russell.
\newblock Actionvlad: Learning spatio-temporal aggregation for action
  classification.
\newblock In {\em Proceedings of the IEEE Conference on Computer Vision and
  Pattern Recognition}, pages 971--980, 2017.

\bibitem{goyal2017something}
Raghav Goyal, Samira~Ebrahimi Kahou, Vincent Michalski, Joanna Materzynska,
  Susanne Westphal, Heuna Kim, Valentin Haenel, Ingo Fruend, Peter Yianilos,
  Moritz Mueller-Freitag, et~al.
\newblock The" something something" video database for learning and evaluating
  visual common sense.
\newblock In {\em ICCV}, volume~1, page~3, 2017.

\bibitem{gu2018ava}
Chunhui Gu, Chen Sun, David~A Ross, Carl Vondrick, Caroline Pantofaru, Yeqing
  Li, Sudheendra Vijayanarasimhan, George Toderici, Susanna Ricco, Rahul
  Sukthankar, et~al.
\newblock Ava: A video dataset of spatio-temporally localized atomic visual
  actions.
\newblock In {\em Proceedings of the IEEE Conference on Computer Vision and
  Pattern Recognition}, pages 6047--6056, 2018.

\bibitem{guo2011multi}
Yuhong Guo and Suicheng Gu.
\newblock Multi-label classification using conditional dependency networks.
\newblock In {\em IJCAI Proceedings-International Joint Conference on
  Artificial Intelligence}, volume~22, page 1300. Citeseer, 2011.

\bibitem{hahnloser2001permitted}
Richard~HR Hahnloser and H~Sebastian Seung.
\newblock Permitted and forbidden sets in symmetric threshold-linear networks.
\newblock In {\em Advances in neural information processing systems}, pages
  217--223, 2001.

\bibitem{he2016deep}
Kaiming He, Xiangyu Zhang, Shaoqing Ren, and Jian Sun.
\newblock Deep residual learning for image recognition.
\newblock In {\em Proceedings of the IEEE conference on computer vision and
  pattern recognition}, pages 770--778, 2016.

\bibitem{horn1981determining}
Berthold~KP Horn and Brian~G Schunck.
\newblock Determining optical flow.
\newblock In {\em Techniques and Applications of Image Understanding}, volume
  281, pages 319--331. International Society for Optics and Photonics, 1981.

\bibitem{hussein2019timeception}
Noureldien Hussein, Efstratios Gavves, and Arnold~WM Smeulders.
\newblock Timeception for complex action recognition.
\newblock In {\em Proceedings of the IEEE Conference on Computer Vision and
  Pattern Recognition}, pages 254--263, 2019.

\bibitem{huynh2020interactive}
Dat Huynh and Ehsan Elhamifar.
\newblock Interactive multi-label cnn learning with partial labels.
\newblock In {\em Proceedings of the IEEE/CVF Conference on Computer Vision and
  Pattern Recognition}, pages 9423--9432, 2020.

\bibitem{ibrahim2016hierarchical}
Mostafa~S Ibrahim, Srikanth Muralidharan, Zhiwei Deng, Arash Vahdat, and Greg
  Mori.
\newblock A hierarchical deep temporal model for group activity recognition.
\newblock In {\em Proceedings of the IEEE Conference on Computer Vision and
  Pattern Recognition}, pages 1971--1980, 2016.

\bibitem{ioffe2015batch}
Sergey Ioffe and Christian Szegedy.
\newblock Batch normalization: Accelerating deep network training by reducing
  internal covariate shift.
\newblock {\em arXiv preprint arXiv:1502.03167}, 2015.

\bibitem{ji2020action}
Jingwei Ji, Ranjay Krishna, Li Fei-Fei, and Juan~Carlos Niebles.
\newblock Action genome: Actions as compositions of spatio-temporal scene
  graphs.
\newblock In {\em Proceedings of the IEEE/CVF Conference on Computer Vision and
  Pattern Recognition}, pages 10236--10247, 2020.

\bibitem{jiang2018human}
Jianwen Jiang, Yu Cao, Lin Song, Shiwei Zhang4~Yunkai Li, Ziyao Xu, Qian Wu,
  Chuang Gan, Chi Zhang, and Gang Yu.
\newblock Human centric spatio-temporal action localization.
\newblock In {\em ActivityNet Workshop on CVPR}, 2018.

\bibitem{kay2017kinetics}
Will Kay, Joao Carreira, Karen Simonyan, Brian Zhang, Chloe Hillier, Sudheendra
  Vijayanarasimhan, Fabio Viola, Tim Green, Trevor Back, Paul Natsev, et~al.
\newblock The kinetics human action video dataset.
\newblock {\em arXiv preprint arXiv:1705.06950}, 2017.

\bibitem{krizhevsky2012imagenet}
Alex Krizhevsky, Ilya Sutskever, and Geoffrey~E Hinton.
\newblock Imagenet classification with deep convolutional neural networks.
\newblock In {\em Advances in neural information processing systems}, pages
  1097--1105, 2012.

\bibitem{kuehne2011hmdb}
Hildegard Kuehne, Hueihan Jhuang, Est{\'\i}baliz Garrote, Tomaso Poggio, and
  Thomas Serre.
\newblock Hmdb: a large video database for human motion recognition.
\newblock In {\em 2011 International Conference on Computer Vision}, pages
  2556--2563. IEEE, 2011.

\bibitem{li2018unified}
Dong Li, Ting Yao, Ling-Yu Duan, Tao Mei, and Yong Rui.
\newblock Unified spatio-temporal attention networks for action recognition in
  videos.
\newblock {\em IEEE Transactions on Multimedia}, 21(2):416--428, 2018.

\bibitem{li2017concurrent}
Xinyu Li, Yanyi Zhang, Jianyu Zhang, Shuhong Chen, Ivan Marsic, Richard~A
  Farneth, and Randall~S Burd.
\newblock Concurrent activity recognition with multimodal cnn-lstm structure.
\newblock {\em arXiv preprint arXiv:1702.01638}, 2017.

\bibitem{meng2019interpretable}
Lili Meng, Bo Zhao, Bo Chang, Gao Huang, Wei Sun, Frederick Tung, and Leonid
  Sigal.
\newblock Interpretable spatio-temporal attention for video action recognition.
\newblock In {\em Proceedings of the IEEE International Conference on Computer
  Vision Workshops}, pages 0--0, 2019.

\bibitem{mosabbeb2014multi}
Ehsan~Adeli Mosabbeb, Ricardo Cabral, Fernando De~la Torre, and Mahmood Fathy.
\newblock Multi-label discriminative weakly-supervised human activity
  recognition and localization.
\newblock In {\em Asian conference on computer vision}, pages 241--258.
  Springer, 2014.

\bibitem{paszke2017automatic}
Adam Paszke, Sam Gross, Soumith Chintala, Gregory Chanan, Edward Yang, Zachary
  DeVito, Zeming Lin, Alban Desmaison, Luca Antiga, and Adam Lerer.
\newblock Automatic differentiation in pytorch.
\newblock 2017.

\bibitem{ray2018scenes}
Jamie Ray, Heng Wang, Du Tran, Yufei Wang, Matt Feiszli, Lorenzo Torresani, and
  Manohar Paluri.
\newblock Scenes-objects-actions: A multi-task, multi-label video dataset.
\newblock In {\em Proceedings of the European Conference on Computer Vision
  (ECCV)}, pages 635--651, 2018.

\bibitem{ren2015faster}
Shaoqing Ren, Kaiming He, Ross Girshick, and Jian Sun.
\newblock Faster r-cnn: Towards real-time object detection with region proposal
  networks.
\newblock In {\em Advances in neural information processing systems}, pages
  91--99, 2015.

\bibitem{ryoo2020assemblenet++}
Michael~S Ryoo, AJ Piergiovanni, Juhana Kangaspunta, and Anelia Angelova.
\newblock Assemblenet++: Assembling modality representations via attention
  connections.
\newblock {\em arXiv preprint arXiv:2008.08072}, 2020.

\bibitem{ryoo2019assemblenet}
Michael~S Ryoo, AJ Piergiovanni, Mingxing Tan, and Anelia Angelova.
\newblock Assemblenet: Searching for multi-stream neural connectivity in video
  architectures.
\newblock {\em arXiv preprint arXiv:1905.13209}, 2019.

\bibitem{shen2017disan}
Tao Shen, Tianyi Zhou, Guodong Long, Jing Jiang, Shirui Pan, and Chengqi Zhang.
\newblock Disan: Directional self-attention network for rnn/cnn-free language
  understanding.
\newblock {\em arXiv preprint arXiv:1709.04696}, 2017.

\bibitem{sigurdsson2016hollywood}
Gunnar~A Sigurdsson, G{\"u}l Varol, Xiaolong Wang, Ali Farhadi, Ivan Laptev,
  and Abhinav Gupta.
\newblock Hollywood in homes: Crowdsourcing data collection for activity
  understanding.
\newblock In {\em European Conference on Computer Vision}, pages 510--526.
  Springer, 2016.

\bibitem{simonyan2014two}
Karen Simonyan and Andrew Zisserman.
\newblock Two-stream convolutional networks for action recognition in videos.
\newblock In {\em Advances in neural information processing systems}, pages
  568--576, 2014.

\bibitem{soomro2012ucf101}
Khurram Soomro, Amir~Roshan Zamir, and Mubarak Shah.
\newblock Ucf101: A dataset of 101 human actions classes from videos in the
  wild.
\newblock {\em arXiv preprint arXiv:1212.0402}, 2012.

\bibitem{sozykin2018multi}
Konstantin Sozykin, Stanislav Protasov, Adil Khan, Rasheed Hussain, and
  Jooyoung Lee.
\newblock Multi-label class-imbalanced action recognition in hockey videos via
  3d convolutional neural networks.
\newblock In {\em 2018 19th IEEE/ACIS International Conference on Software
  Engineering, Artificial Intelligence, Networking and Parallel/Distributed
  Computing (SNPD)}, pages 146--151. IEEE, 2018.

\bibitem{srivastava2014dropout}
Nitish Srivastava, Geoffrey Hinton, Alex Krizhevsky, Ilya Sutskever, and Ruslan
  Salakhutdinov.
\newblock Dropout: a simple way to prevent neural networks from overfitting.
\newblock {\em The journal of machine learning research}, 15(1):1929--1958,
  2014.

\bibitem{szegedy2015going}
Christian Szegedy, Wei Liu, Yangqing Jia, Pierre Sermanet, Scott Reed, Dragomir
  Anguelov, Dumitru Erhan, Vincent Vanhoucke, and Andrew Rabinovich.
\newblock Going deeper with convolutions.
\newblock In {\em Proceedings of the IEEE conference on computer vision and
  pattern recognition}, pages 1--9, 2015.

\bibitem{tran2015learning}
Du Tran, Lubomir Bourdev, Rob Fergus, Lorenzo Torresani, and Manohar Paluri.
\newblock Learning spatiotemporal features with 3d convolutional networks.
\newblock In {\em Proceedings of the IEEE international conference on computer
  vision}, pages 4489--4497, 2015.

\bibitem{tran2019video}
Du Tran, Heng Wang, Lorenzo Torresani, and Matt Feiszli.
\newblock Video classification with channel-separated convolutional networks.
\newblock In {\em Proceedings of the IEEE International Conference on Computer
  Vision}, pages 5552--5561, 2019.

\bibitem{vaswani2017attention}
Ashish Vaswani, Noam Shazeer, Niki Parmar, Jakob Uszkoreit, Llion Jones,
  Aidan~N Gomez, {\L}ukasz Kaiser, and Illia Polosukhin.
\newblock Attention is all you need.
\newblock In {\em Advances in neural information processing systems}, pages
  5998--6008, 2017.

\bibitem{wang2016cnn}
Jiang Wang, Yi Yang, Junhua Mao, Zhiheng Huang, Chang Huang, and Wei Xu.
\newblock Cnn-rnn: A unified framework for multi-label image classification.
\newblock In {\em Proceedings of the IEEE conference on computer vision and
  pattern recognition}, pages 2285--2294, 2016.

\bibitem{wang2019hallucinating}
Lei Wang, Piotr Koniusz, and Du Huynh.
\newblock Hallucinating idt descriptors and i3d optical flow features for
  action recognition with cnns.
\newblock In {\em Proceedings of the 2019 International Conference on Computer
  Vision}. IEEE, Institute of Electrical and Electronics Engineers, 2019.

\bibitem{wang2016temporal}
Limin Wang, Yuanjun Xiong, Zhe Wang, Yu Qiao, Dahua Lin, Xiaoou Tang, and Luc
  Van~Gool.
\newblock Temporal segment networks: Towards good practices for deep action
  recognition.
\newblock In {\em European conference on computer vision}, pages 20--36.
  Springer, 2016.

\bibitem{wang2020multi}
Qian Wang and Ke Chen.
\newblock Multi-label zero-shot human action recognition via joint latent
  ranking embedding.
\newblock {\em Neural Networks}, 122:1--23, 2020.

\bibitem{wang2018non}
Xiaolong Wang, Ross Girshick, Abhinav Gupta, and Kaiming He.
\newblock Non-local neural networks.
\newblock In {\em Proceedings of the IEEE Conference on Computer Vision and
  Pattern Recognition}, pages 7794--7803, 2018.

\bibitem{wang2018videos}
Xiaolong Wang and Abhinav Gupta.
\newblock Videos as space-time region graphs.
\newblock In {\em Proceedings of the European Conference on Computer Vision
  (ECCV)}, pages 399--417, 2018.

\bibitem{wu2019long}
Chao-Yuan Wu, Christoph Feichtenhofer, Haoqi Fan, Kaiming He, Philipp
  Krahenbuhl, and Ross Girshick.
\newblock Long-term feature banks for detailed video understanding.
\newblock In {\em Proceedings of the IEEE Conference on Computer Vision and
  Pattern Recognition}, pages 284--293, 2019.

\bibitem{wu2020multigrid}
Chao-Yuan Wu, Ross Girshick, Kaiming He, Christoph Feichtenhofer, and Philipp
  Krahenbuhl.
\newblock A multigrid method for efficiently training video models.
\newblock In {\em Proceedings of the IEEE/CVF Conference on Computer Vision and
  Pattern Recognition}, pages 153--162, 2020.

\bibitem{xue2011correlative}
Xiangyang Xue, Wei Zhang, Jie Zhang, Bin Wu, Jianping Fan, and Yao Lu.
\newblock Correlative multi-label multi-instance image annotation.
\newblock In {\em 2011 International Conference on Computer Vision}, pages
  651--658. IEEE, 2011.

\bibitem{yazici2020orderless}
Vacit~Oguz Yazici, Abel Gonzalez-Garcia, Arnau Ramisa, Bartlomiej Twardowski,
  and Joost van~de Weijer.
\newblock Orderless recurrent models for multi-label classification.
\newblock In {\em Proceedings of the IEEE/CVF Conference on Computer Vision and
  Pattern Recognition}, pages 13440--13449, 2020.

\bibitem{yeung2018every}
Serena Yeung, Olga Russakovsky, Ning Jin, Mykhaylo Andriluka, Greg Mori, and Li
  Fei-Fei.
\newblock Every moment counts: Dense detailed labeling of actions in complex
  videos.
\newblock {\em International Journal of Computer Vision}, 126(2-4):375--389,
  2018.

\bibitem{zhou2018temporal}
Bolei Zhou, Alex Andonian, Aude Oliva, and Antonio Torralba.
\newblock Temporal relational reasoning in videos.
\newblock In {\em Proceedings of the European Conference on Computer Vision
  (ECCV)}, pages 803--818, 2018.

\end{thebibliography}
}

\end{document}